\documentclass[twocolumn,conference]{IEEEtran}
\usepackage[T1]{fontenc}

\usepackage{xcolor}
\usepackage{amsbsy}
\usepackage{amssymb}
\usepackage{graphicx}

\ifCLASSINFOpdf

\else

\fi

\usepackage{multirow}
\usepackage{float}
\usepackage{hyperref}
\usepackage[export]{adjustbox}
\usepackage{xcolor}
\usepackage{caption}
\usepackage{subcaption}
\usepackage{afterpage}
\usepackage[labelsep=period, figurename=Figure]{caption}

\usepackage{amsmath}
\usepackage{verbatim}
\usepackage{graphicx}

\usepackage{tikz}
\usepackage{textcomp}
\usepackage{hyperref}
\usepackage{lipsum}
\usepackage{comment}
\usepackage{algpseudocode}
\usepackage{amsmath}
\usepackage{algorithm}
\usepackage{cite}
\usepackage{graphicx}
\usepackage{textcomp}
\usepackage{xcolor}
\usepackage{amsmath,amssymb,amsfonts}
\usepackage{hyperref}
\usepackage{booktabs}

\usepackage{enumitem}
\setlist[enumerate,1]{label={\arabic*.}}
\makeatother

\begin{document}
\addtolength{\textfloatsep}{-2.19pt}

%

\title{Large Language Model-Driven Curriculum Design for Mobile Networks}


\author{
    \IEEEauthorblockN{Omar Erak$^{1}$, Omar Alhussein$^{1}$, Shimaa Naser$^{2}$, Nouf Alabbasi$^{1}$, De Mi$^{3}$, Sami Muhaidat$^{2,4}$}
    \IEEEauthorblockA{$^{1}$KU 6G Research Centre, Department of Computer Science, Khalifa University, Abu Dhabi, UAE\\
                      $^{2}$KU 6G Research Centre, Department of Computer and Information Engineering, Khalifa University, Abu Dhabi, UAE\\
                      $^{3}$Future Information Networks Research Cluster, College of Computing, Birmingham City University, UK\\
                      $^{4}$Department of Systems and Computer Engineering, Carleton University, Ottawa, Canada\\
                      Emails: 100065156@ku.ac.ae, omar.alhussein@ku.ac.ae, shimaa.naser@ku.ac.ae, 100064507@ku.ac.ae, \\de.mi@bcu.ac.uk, muhaidat@ieee.org}
}

\maketitle

\begin{abstract}

This study introduces an innovative framework that employs large language models (LLMs) to automate the design and generation of curricula for reinforcement learning (RL). As mobile networks evolve towards the 6G era, managing their increasing complexity and dynamic nature poses significant challenges. Conventional RL approaches often suffer from slow convergence and poor generalization due to conflicting objectives and the large state and action spaces associated with mobile networks. To address these shortcomings, we introduce curriculum learning, a method that systematically exposes the RL agent to progressively challenging tasks, improving convergence and generalization.
However, curriculum design typically requires extensive domain knowledge and manual human effort. Our framework mitigates this by utilizing the generative capabilities of LLMs to automate the curriculum design process, significantly reducing human effort while improving the RL agent's convergence and performance. We deploy our approach within a simulated mobile network environment and demonstrate improved RL convergence rates, generalization to unseen scenarios, and overall performance enhancements.
As a case study, we consider autonomous coordination and user association in mobile networks. Our obtained results highlight the potential of combining LLM-based curriculum generation with RL for managing next-generation wireless networks, marking a significant step towards fully autonomous network operations. 

\end{abstract}



\IEEEpeerreviewmaketitle

\section{Introduction}
As we move beyond the fifth-generation (5G) networks towards 6G networks, we are stepping into an era characterized by remarkable data rates, bandwidth, latency, and reliability.  These improvements aim to enhance users' quality of experience (QoE) and meet the growing demand for mobile connectivity and services. The integration of sensing capabilities within access networks is anticipated to further blend the physical and virtual worlds. Thanks to advancements in physical layer technology and due to network virtualization, 6G is anticipated to accommodate numerous novel use cases and business models beyond what is conceivable by current networks. These encompass extended reality applications, autonomous vehicles, and Internet of Things ecosystems \cite{6GVision}.

As these networks evolve, the complexity of management and coordination while ensuring optimal performance amid dynamically changing environments becomes increasingly challenging. Traditional network management methods often struggle to keep pace with the network's dynamics, leading to sub-optimal performance and reliability issues \cite{UserAsscSurvery}.  
Thus, to unlock the full potential of 6G, it is vital to develop self-optimizing networks capable of intelligent decision-making to reduce the need for frequent human intervention. This evolution is essential for achieving optimal connectivity, personalized service delivery, and seamless automation \cite{metanet}.



Reinforcement Learning (RL) offers a promising role in tackling this challenge, thanks to its ability to learn and adapt to dynamic environments. By continuously interacting with the environment, RL algorithms can develop strategies that maximize long-term rewards, making them particularly suitable for managing the complex dynamics of mobile networks \cite{9667095}. However, a major hurdle in the practical application of RL to such tasks is the issue of poor convergence and generalization, especially in scenarios with large state and action spaces, as observed in network coordination, user association, and resource management tasks \cite{RLMobile}. 

To this end, curriculum learning (CL) has been proposed as a method to overcome the aforementioned challenges. By structurally introducing the RL agent to the problem domain, starting from simpler scenarios and progressively moving to more complex ones, CL can significantly enhance the learning efficiency, performance, and generalization capability of RL algorithms \cite{CLsurvey}. This approach mimics the human learning process, where foundational knowledge is built first, followed by more complex concepts, instead of directly attempting to tackle a complex concept from scratch. 

Nevertheless, designing an effective curriculum suitable for training RL agents is deeply reliant on domain expertise and is often a labor-intensive process. Predicated on the fact that state-of-the-art large language models (LLMs) are trained on extensive corpora of knowledge, and inspired by their emergent capabilities such as reasoning and generalization, we investigate leveraging these models to automate the generation of curriculum and associated rewards. By automating curriculum design with LLMs, we significantly reduce manual effort, accelerating the development of efficient RL agents. This approach not only makes the deployment of advanced RL agents for 6G networks faster and more accessible but also paves the way for fully autonomous networks with minimal human intervention.


In this paper, as a case study, we consider user association in mobile networks. We address the unique challenges associated with RL in this domain, elaborate on the principles and advantages of CL, and illustrate how LLMs offer a pathway to streamline and automate curriculum design. Moreover, our research provides a thorough evaluation of the impact of LLM-driven curriculum generation on RL convergence, the agent's generalization capability in unseen environments, and the overall performance within simulated environments. Results demonstrate significant improvements in model convergence, adaptability, and generalization.
To the best of our knowledge, this is the first work to systematically employ and evaluate LLMs for curriculum generation in RL, and the first to apply it to mobile networks. 

The rest of this paper is organized as follows. In Section II, we provide an overview of related works in RL for mobile networks, as well as a general literature review on CL. Section III introduces the system model and problem statement. Section IV presents the DRL approach without a curriculum, and Section V presents the DRL approach with an LLM-based curriculum. In Section VI, we outline our evaluation framework and provide a thorough discussion of the results and their implications. Finally, Section VII concludes the paper with a discussion on the importance of our results and future work.

\section{Related Work}

\subsection{RL for Mobile Networks}
 RL has been extensively explored in mobile networks emphasizing the potential of RL agents to adapt to dynamic environments. In \cite{sana2020multi}, the authors propose a multi-agent RL framework for adaptive user association in dynamic mmWave networks. The proposed approach adapts better to changing dynamics and outperforms conventional methods that do not rely on RL. This highlights the potential of deploying RL agents in dynamic network environments. 
On the other hand, in \cite{schneider2023deepcomp}, Schneider et al. leveraged deep reinforcement learning (DRL) for optimizing Coordinated Multipoint (CoMP) in mobile networks. CoMP is crucial for enhancing network capacity by enabling users to connect to multiple cells simultaneously. Traditional approaches to CoMP optimization often rely on detailed system knowledge or heuristic algorithms, which may not always be practical or effective given the dynamic nature of mobile networks. The proposed methods by \cite{schneider2023deepcomp} are designed to adapt to various network conditions, user movements, and resource allocation schemes, showing the potential to significantly improve QoE over traditional methods. 

The methods discussed above highlight the successes of RL in optimizing and advancing mobile networks. 
However, despite these advances, the long convergence times and poor generalization inherent to DRL pose notable challenges,  potentially hindering its immediate application in rapidly changing network environments. Furthermore, none of the methods utilize CL in their frameworks, and specifically, LLMs for automated CL in this context remain unexplored.

\subsection{CL and Automated Curriculum Generation}
Curriculum learning, which involves the strategic ordering of training samples to speed up the learning process, has been recognized for its potential to improve the efficiency of RL algorithms. Bengio et al. first introduced this concept, illustrating how it can mimic human learning processes to benefit machine learning models \cite{bengio2009curriculum}. Since then, CL has gained a lot of research interest from the community \cite{AwesomeCL}. It is worth noting that CL has proven to be successful in different tasks such as supervised and RL problems. For instance, Graves et al. demonstrated using a stochastic syllabus (by sampling from a set of human-generated tasks) and a progress signal resulting in accelerated learning and improved performance in a supervised learning context \cite{graves2017automated}. In the context of RL, Narvekar et al. introduced a novel approach to CL, focusing on autonomous task sequencing for customized curriculum design. Their work highlights the potential of dynamically generated curricula to enhance learning speed and adaptability in RL paving the way for more personalized and efficient learning strategies in complex environments \cite{narvekar2017autonomous}. In \cite{du2022it}, Yuqing et al. propose automatic curriculum generation by using four different agents to aid with the curriculum generation.

CL is a powerful strategy for both RL and supervised learning domains. In RL specifically, the gradual introduction of tasks tailored to the learner's growing capabilities has been shown to accelerate convergence and improve generalization. Recent research has explored automating the generation of these curricula, aiming to optimize learning paths based on the learner's interactions and progress, further reducing the need for manual curriculum design as discussed above. Despite these advancements, leveraging LLMs to aid in CL remains largely uncharted, and a complete framework that integrates LLM curriculum generation within the RL agent training phase has not been explored or proposed yet.

\section{System Model and Problem Statement}


\subsection{System Model} 
We consider a mobile network environment which consists of $M$ User Equipment (UEs) $u_i \in \cal{U}$ and $N$ base stations $b_j \in \cal{B}$, where $M=|\cal{U}|$ and $N=|\cal{B}|$ are customizable parameters. The transmission of base stations is synchronized and transmission occurs in discrete time steps $t=1,2,..,T$. It is assumed that the mobility model for each UE $u_i$ follows the random waypoint model \cite{hyytia2007random}, and the velocity of each user is customized within the environment. Each user-station pair has an associated data rate at time $t$ which is denoted by $D_{ij}(t)$. The SINR, denoted as \(\rho_{ij}(t)\), is the measured \(\rho\) from base station \(b_j\) to user equipment \(u_i\) at time \(t\). UE \(u_i\) can only connect to base station \(b_j\) if \(\rho_{ij}(t)\) is above a specific threshold. The Hata-Okumura propagation model is used for predicting the path loss in urban areas \cite{Okumura}.

\subsection{Problem Statement}
\begin{figure}[ht]
    \centering
    \includegraphics[width=0.4\textwidth]{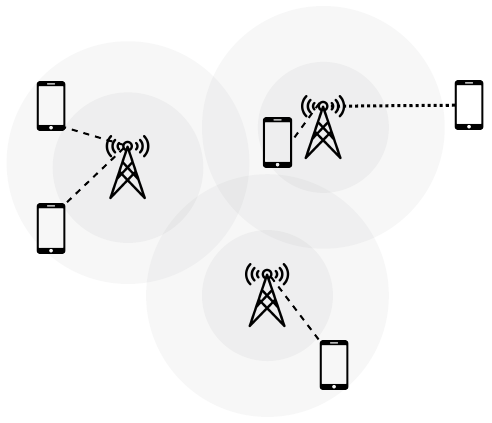}
    
    \caption{Example environment with $M = 5$, and $N = 3$. dashed lines represent connections, while the dotted line represents a failure to meet $\rho_{min}$.}
    \label{fig: Env}
\end{figure}

The primary objective of our problem is to maximize the long-term average QoE across all UEs in a mobile wireless network environment by determining the optimal connectivity configurations between UEs and BSs across discrete time intervals \(t\). To quantify the system's performance we adopt a specific measure of QoE that accounts for the quality of the connection across UEs. The QoE for each UE $i$ when connected to base station $j$ is formulated as a utility function $Q_{ij}(t)$ that maps the logarithm of the received data rate $D_{ij}(t)$ to a normalized scale ranging from $0$ to $1$. This is given by:

\begin{equation}
    Q_{ij}(t) = 
\begin{cases} 
\frac{\log(D_{ij}(t)) - \log(D_{\min})}{\log(D_{\max}) - \log(D_{\min})}, & \text{if } D_{ij}(t) > 0 \\
0, & \text{if } D_{ij}(t) = 0
\end{cases}
\end{equation}

\noindent where \( D_{ij}(t) \) is the data rate received by user \( i \) from station \( j \) at time \( t \). Parameters \( D_{\min} \) and \( D_{\max} \) are the minimum and maximum data rates observed across all user-station pairs, respectively.
%
%
%
%
%
%
The mathematical formulation of the objective and constraints is as follows:
\begin{subequations}
\begin{align}
& \hspace{5pt} \max_{x_{ij}} \quad \lim_{{T \to \infty}} \frac{1}{T} \sum_{{t \in T}} \frac{1}{M} \sum_{{i=1}}^{M} \sum_{{j=1}}^{N} x_{ij}(t)Q_{ij}(t) \nonumber \tag{P1}  \\
& \text{subject to:} \nonumber \\
\quad & x_{ij}(t) \in \{0, 1\}, \;\; \quad \forall i \in \{1, 2, ..., M\},\forall j \in \{1, 2, ..., N\} \label{eq:binary_constraint} \tag {P1.a} \\
& \sum_{{j=1}}^{N} x_{ij}(t) \leq N, \;\; \forall i \in \{1, 2, ..., M\} \label{eq:connection_constraint} \tag {P1.b} \\
 &\rho_{ij} \geq \rho_{\text{min}},\;\;  \forall i \in \{1, 2, ..., M\}, \forall j \in \{1,2,\dots,N\} \label{eq:sinr_constraint}\tag {P1.c}
\end{align}
\end{subequations}

\noindent where $x_{ij}(t) \in \{0,1\}$ is a binary decision variable indicating whether user $i$ is connected to station $j$ at time $t$. It takes a value of 1 if there is a connection, and 0 otherwise. Parameter \( \rho_{\text{min}} \) is the minimum acceptable SINR for maintaining a connection.
%
%
Constraint (\ref{eq:binary_constraint}) ensures that the decision variable \(x_{ij}(t)\) can only take binary values, indicating a connection status between UEs and BSs of either connected (1) or not connected (0). Constraint (\ref{eq:connection_constraint}) stipulates that the sum of connections for each UE to all BSs at any given time 
$t$ does not exceed $N$. This allows any UE to connect with multiple BSs simultaneously.  Constraint (\ref{eq:sinr_constraint}) requires $\rho_{ij}$ to be above a minimum threshold for a connection to be established. Fig. \ref{fig: Env} depicts an example environment where some UEs are connected and others are not.

\section{DRL Without Curriculum Design}  \label{sec:nocl}

\subsection{DRL Agent}
The DRL agent operates within the framework of a Markov Decision Process (MDP), characterized by a tuple $(\mathcal{S}, \mathcal{A}, \mathcal{P}, \mathcal{R})$, where $\mathcal{S}$ represents the states, $\mathcal{A}$ denotes the actions, $\mathcal{P}$ signifies the transition probabilities of the environment dynamics, and $\mathcal{R}$ is the reward function.
The learning objective for the agent is to find an optimal policy that maximizes the expected return $K(t)$ which is defined as the discounted sum of future rewards:

\begin{equation}
    K(t) = \sum_{k=0}^{\infty} \gamma^k R(t+k+1),
\end{equation}
where $\gamma$ denotes the discount factor.

For our RL agent, we adopt the proximal policy optimization (PPO) algorithm which enhances the efficiency and stability of the agent's learning process by optimizing a surrogate objective function \cite{schulman2017proximal}. This function constrains the policy change in a small range using a clip, which helps facilitate smoother policy convergence. The objective function for the PPO algorithm is given by:

\begin{equation}
    L^{CLIP}(\theta) = \hat{\mathbb{E}}_t \left[ \min\left(r_t(\theta) \hat{A}_t, \text{clip}\left(r_t(\theta), 1 - \epsilon, 1 + \epsilon\right) \hat{A}_t\right) \right]
\end{equation}

\noindent where $\hat{A}_t$ represents the estimated advantage at time $t$, $r_t(\theta)$ is the probability ratio $\frac{\pi_\theta(a_t|s_t)}{\pi_{\theta_{old}}(a_t|s_t)}$, and $\epsilon$ is a hyperparameter that defines the clipping range to maintain the policy update within a designated region.

\subsection{States, Actions, and Rewards} \label{subsec:SAR}
At each timestep $t$, the action space $\mathcal{A}$ consists of decision variables $x_{ij}(t),\,i=\{1,2,\dots,M \}, j=\{1,2,\dots,N \}$, which determine the connectivity status between \(u_i\) and  \(b_j\) at each timestep. The agent's state space $\mathcal{S}$ at each timestep $t$ is specified by: \(x_{ij}(t)\), \(\rho_{ij}(t)\), and \(Q_{ij}(t)\), $i=\{1,2,\dots,M \}$, $j=\{1,2,\dots,N \}$. The agent's reward at each timestep $t$ is the average QoE over all UEs, expressed as
\begin{equation}
    R(t) = \frac{1}{M} \sum_{{i=1}}^{M} \sum_{{j=1}}^{N} Q_{ij}(t). \label{eq:R(t)}
\end{equation}
The sizes of the action and observation states depend on $M$ and $N$ which could vary over time. However, these sizes must be fixed as they are used as inputs and outputs for the neural network. Therefore, we define $M_{max}$ as the maximum number of supported UEs and $N_{max}$ as the maximum number of BSs. If $M < M_{max}$ or $N < N_{max}$, the unused states are padded with zeros.

\section{DRL with LLM-based Curriculum Design}

\subsection{The Framework}
\begin{figure}[ht]
    \centering
    \includegraphics[width=0.45\textwidth, height = 8cm]{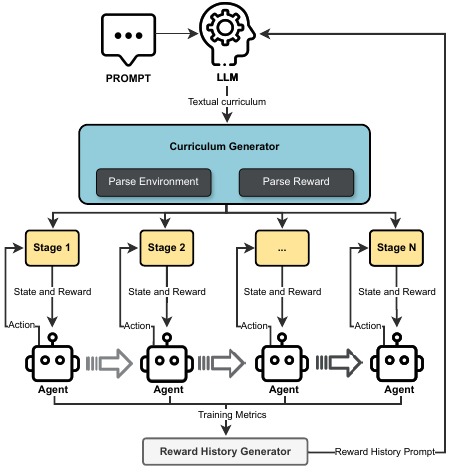}
    \caption{
Framework for LLM generated curriculum learning.}
    \label{fig:framework}
\end{figure}

In this paper, we leverage the capabilities of LLMs for the dynamic generation and adaptation of curricula to aid an RL agent in accomplishing a complex task. As depicted in Fig. \ref{fig:framework}, the process begins with entering a prompt to an LLM. The prompt describes the target task, including the observation and action spaces of the DRL agent. The LLM is also asked to produce a curriculum and corresponding reward functions in a specific format. Accordingly, the LLM generates a curriculum that is divided into multiple stages, each tailored to progressively challenge the RL agent as it improves its skills. Each stage is characterized by distinct objectives and difficulty levels. For example, early stages may focus on fundamental skills and simple decision-making scenarios with a low number of users. As the agent demonstrates strong proficiency in the foundational aspects of the problem, the curriculum escalates to more complex scenarios.

\begin{figure}[ht]
    \centering
    \includegraphics[width=0.5\textwidth, height = 6.4cm]{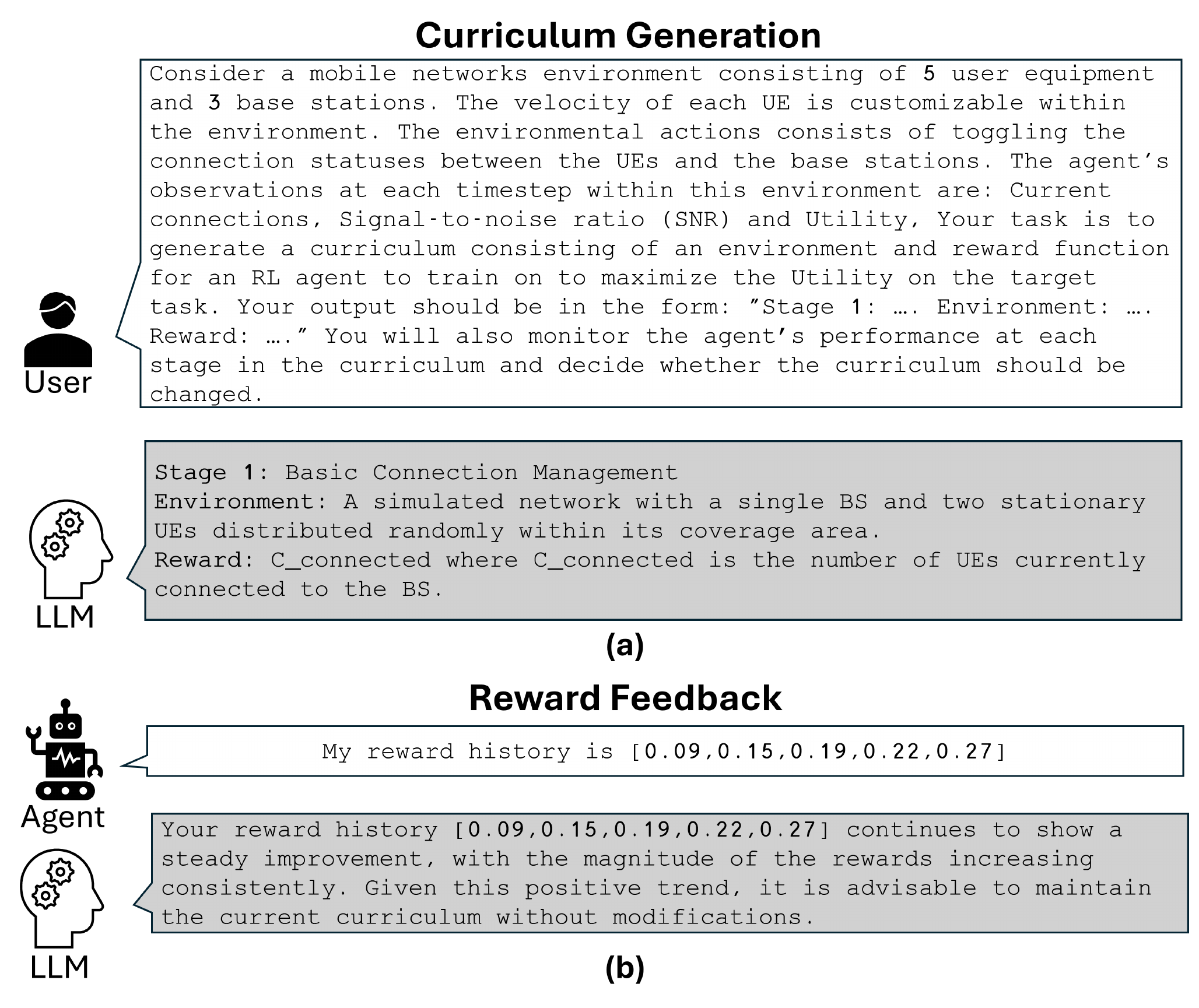}
    \caption{Example prompt followed by LLM-generated stage 1 of the curriculum. The second prompt is the agent's reward history followed by the LLM's decision on curriculum changes if needed.}
    \label{fig:Prompt}
\end{figure}

An example of both the input prompt and the generated curriculum for stage 1 is shown in Fig. \ref{fig:Prompt}(a). The agent undergoes training with the curriculum and progresses to subsequent stages. To monitor the progress, the agent's training metrics are captured by a reward history generator that tracks rewards at each step. As shown in Fig. \ref{fig:Prompt}(b), the data is used to generate a new text prompt for the LLM to determine if and how the curriculum should be adapted.

As outlined in Algorithm \ref{alg:algorithm1}, the process begins with the initialization of the environment ($\mathbb{E}$), the target task ($\mathbb{T}$), the LLM ($L$), and the RL agent ($A$). The first step involves generating a textual-tailored curriculum (\textit{C}) based on a textual description of the environment and the target task (\textit{Desc}) \textit{(line 4)}. The textual description serves as a prompt for the LLM to generate the output (\textit{C}) \textit{(line 5)}. This output, (\textit{C}), is then parsed and a list of curriculum stages and rewards is obtained \textit{(line 6)}. Each stage is characterized by a specific task and environment ($T_s$), along with the corresponding rewards ($R_s$). All of which are generated by the LLM to challenge and progress the RL agent's learning capabilities. The agent progresses to the next stage of the curriculum based on its performance relative to a predefined threshold (\textit{$\theta$}) \textit{(lines 10-11)}.

An important step in the algorithm is incorporating a feedback loop that enables real-time curriculum adjustment based on the agent's reward history. The reward history is parsed into text and used as a prompt for the LLM \textit{(line 15)}. This adaptive approach ensures that the curriculum in use is helping the agent with its learning and allows the LLM to change the curriculum if the agent is not converging, or progressing too slowly. Upon reaching or exceeding the performance threshold the agent progresses to the following stage iterating through the curriculum until it has been successfully trained on the target task.

\begin{algorithm}
\caption{LLM-Based Curriculum Generation for RL}
\label{alg:algorithm1}
\begin{algorithmic}[1]
\State \textbf{Input:} Environment $\mathbb{E}$; Task $\mathbb{T}$; LLM $L$; Agent $A$
\State \textbf{Output:} $A$ trained on $T$
\State \textbf{Init:} $C \gets \text{""}$, $\theta$, $s \gets 0$, $RH \gets []$, $TR \gets []$
\State $Desc \gets \Call{Describe}{\mathbb{E}, \mathbb{T}}$
\State $C \gets L.\Call{GenTextCur}{Desc}$
\State $TR \gets \Call{ParseAndGenCurr}{C}$
\While{$s < |TR|$}
\State $(T_s, R_s) \gets TR[s]$
\State $RH[s] \gets A.\Call{Train}{T_s, R_s}$
\If{$RH[s] \geq \theta$}
\State $s \gets s + 1$
\EndIf
\If{$\neg$ $\Call{L.CheckRewardIncrease}{RH}$}
\State $C \gets L.\Call{AdjCurText}{C, s}$
\State $TR \gets \Call{ParseAndGenCurr}{C}$ 
\State $s \gets \max(s-1, 0)$
\EndIf
\EndWhile
\State \Return $A$
\end{algorithmic}
\end{algorithm}


\subsection{LLM-Generated Curriculum}
Based on the prompt in Fig. \ref{fig:Prompt}, the LLM generates a curriculum that consists of four stages with successively increasing levels of complexity in terms of the task and the environment.



\noindent\textbf{Stage 1 - Basic Connectivity}: The objective is to teach the RL agent to establish basic connectivity between UEs and the nearest BS. The environment is reduced to a single BS and two stationary UEs distributed randomly within its coverage area. That is, the state space $\mathcal{S}^{(S1)}$ consists of variables: \(x_{i1}(t)\), \(\rho_{i1}(t)\), and \(Q_{i1}(t)\), and the action space $\mathcal{A}^{(S1)}$ is \(x_{i1}(t)\) for $i=\{1,2\}$.
The reward function for this stage becomes
\begin{equation}
    R^{(S1)}(t) = \sum_{{i=1}}^{2} x_{i1}(t).
\end{equation}

\noindent\textbf{Stage 2 - Mobility Management}:
Here, the LLM introduces the random waypoint mobility model to the environment, albeit at slow velocities. As the UEs move slowly, the agent needs to learn to maintain connectivity. The new environment is set to contain two BSs and two UEs. That is, the state space $\mathcal{S}^{(S2)}$ consists of variables: \(x_{ij}(t)\), \(\rho_{ij}(t)\), and \(Q_{ij}(t)\), and the action space $\mathcal{A}^{(S2)}$ is \(x_{ij}(t)\) for $i=\{1,2\}$ and $j=\{1,2\}$.
For this stage, the reward function is

\begin{equation}
    R^{(S2)}(t) =  \sum_{i=1}^{2} \sum_{j=1}^{2} (x_{ij}(t)x_{ij}(t-1)).
\end{equation}


\noindent\textbf{Stage 3 - Preliminary QoE Maximization}: In this stage, the agent is now being prepared for the final target task by focusing on maximizing the QoE for all UEs in a simplified network setup with two BSs and three  UEs. The random waypoint model is used, and UE velocities are completely randomized. This stage serves as a scaled-down version of the final target task. The state space $\mathcal{S}^{(S3)}$ consists of variables: \(x_{ij}(t)\), \(\rho_{ij}(t)\), and \(Q_{ij}(t)\), and the action space $\mathcal{A}^{(S3)}$ is \(x_{ij}(t)\) for $i=\{1,2,3\}$ and $j=\{1,2\}$.  The generated reward function is given by:
\begin{equation}
    R^{(S3)}(t) = \frac{1}{3} \sum_{{i=1}}^{3} \sum_{{j=1}}^{2} Q_{ij}(t).
\end{equation}

\noindent \textbf{Stage 4 - Final Target Task}: The final stage is equivalent to the intended final environment with the states, actions and rewards are as specified in subsection \ref{subsec:SAR} for $M = 5$, and $N=3$.


\section{Experimental Results and Analysis}
In this paper, following the system model, we utilize the open-source simulation environment developed by Schneider et al. \cite{mobileenv}. For reproducibility and reuse, our source code is made publicly available \cite{repo}. 
For the LLM, we utilize GPT-4 by using OpenAI's publicly available application programming interfaces (APIs) \cite{openai2024gpt4}. We employ a feed-forward neural network architecture with two hidden layers, each consisting of 64 neurons. The learning rate and batch size are set to $0.0005$ and $64$, respectively. To assess the efficacy of our proposed LLM-based curriculum design approach, we compare it against the DRL approach implemented without a curriculum.

\begin{figure}[ht]
    \centering
    \includegraphics[width=0.47\textwidth]{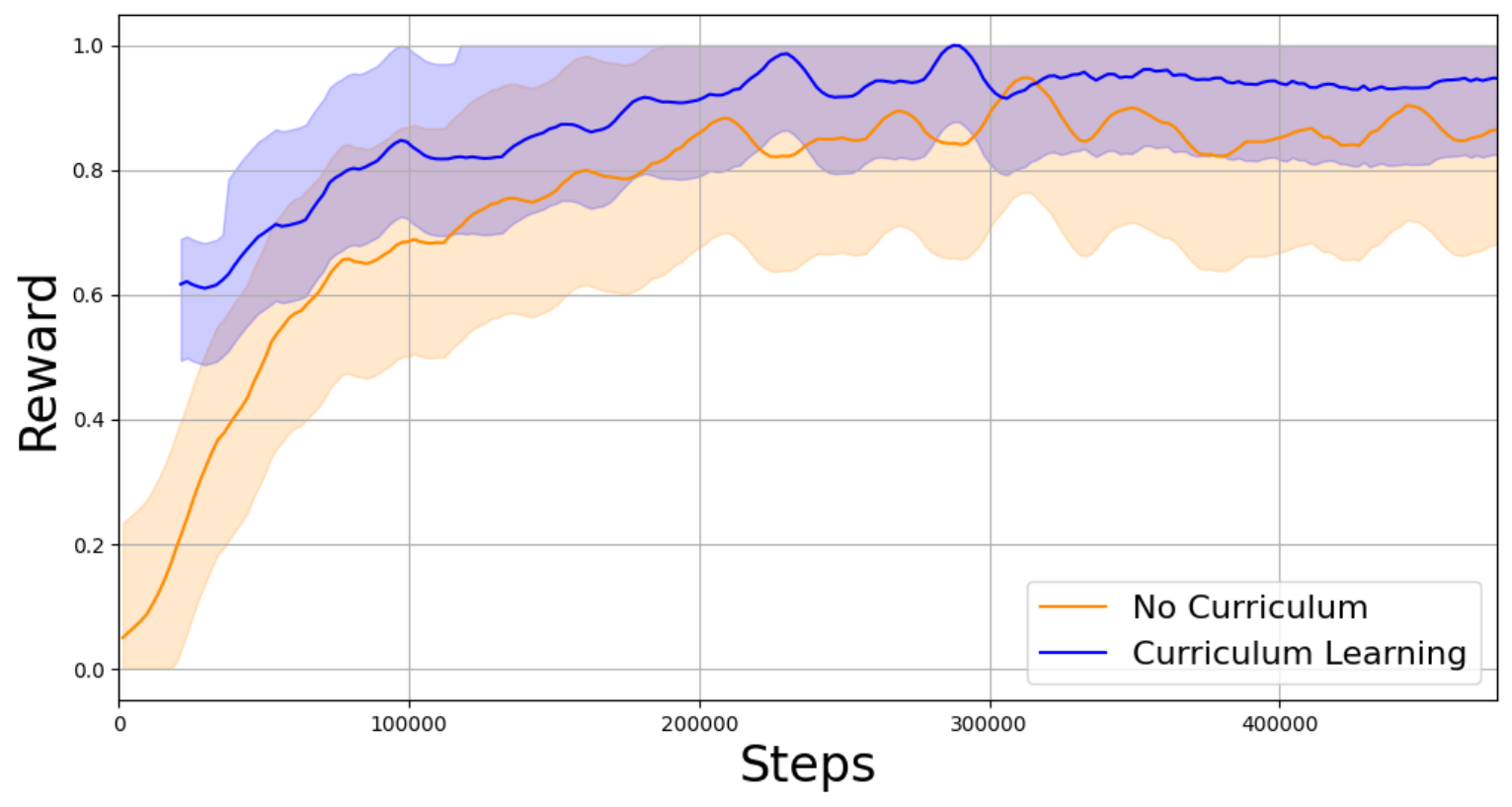}
    \caption{Training performance of the curriculum-based learning approach and the baseline approach.}
    \label{fig:graph1}
\end{figure}

\noindent \textbf{Faster Convergence with Curriculum Learning}.
As illustrated in Fig. \ref{fig:graph1}, the proposed LLM-based curriculum approach converged in fewer steps than the baseline approach that did not utilize curriculum learning. More specifically, the proposed approach converged approximately 75,000 steps earlier than the baseline method. Notably, the figure shows the curriculum learning approach starting at a delayed episode to account for the episodes spent on the early stages of the curriculum. We also note that the curriculum-based approach experiences less fluctuations and maintains a higher reward throughout the training episodes. This result demonstrates the effectiveness of the LLM-generated curriculum in accelerating the learning process, achieving higher rewards, and supports the benefits of structured learning paths in improving convergence speed.

\vspace{0.3cm}


\noindent \textbf{Generalization to Unseen Environments}.
Curriculum learning is often associated with stronger generalization and robustness to unseen environments. To evaluate this, we conducted tests on the trained agents in environments with variations beyond their training conditions, specifically in the number of UEs. Although both approaches were originally trained with up to 5 UEs, here we test the agents with 6 to 10 UEs and also vary the locations of the BSs.

\begin{figure}[ht]
    \centering
    \includegraphics[width=0.47\textwidth]{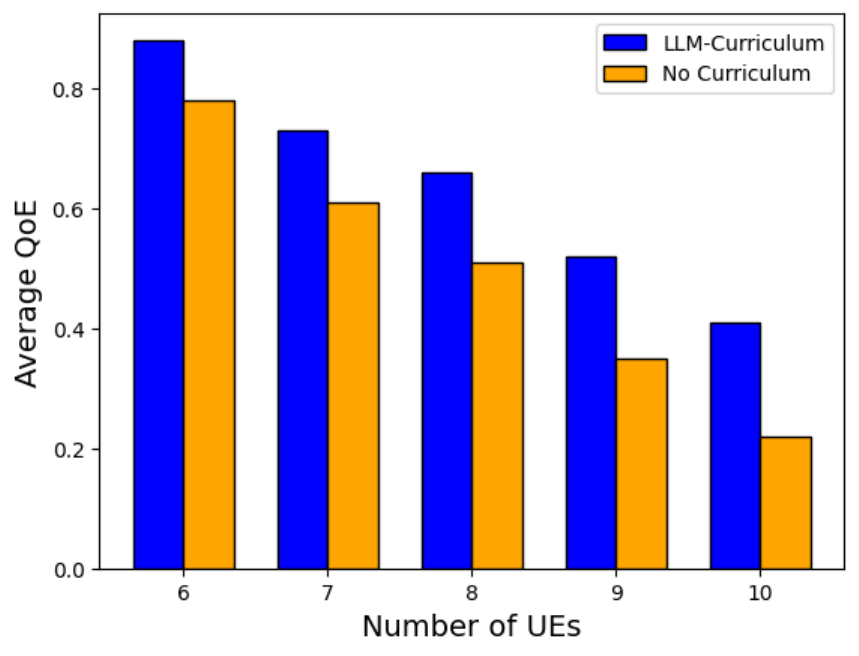}
    \caption{Performance comparison in unseen environments with varying number of UEs.}
    \label{fig:result2}
\end{figure}

\begin{figure}[ht]
    \centering
    \includegraphics[width=0.47\textwidth]{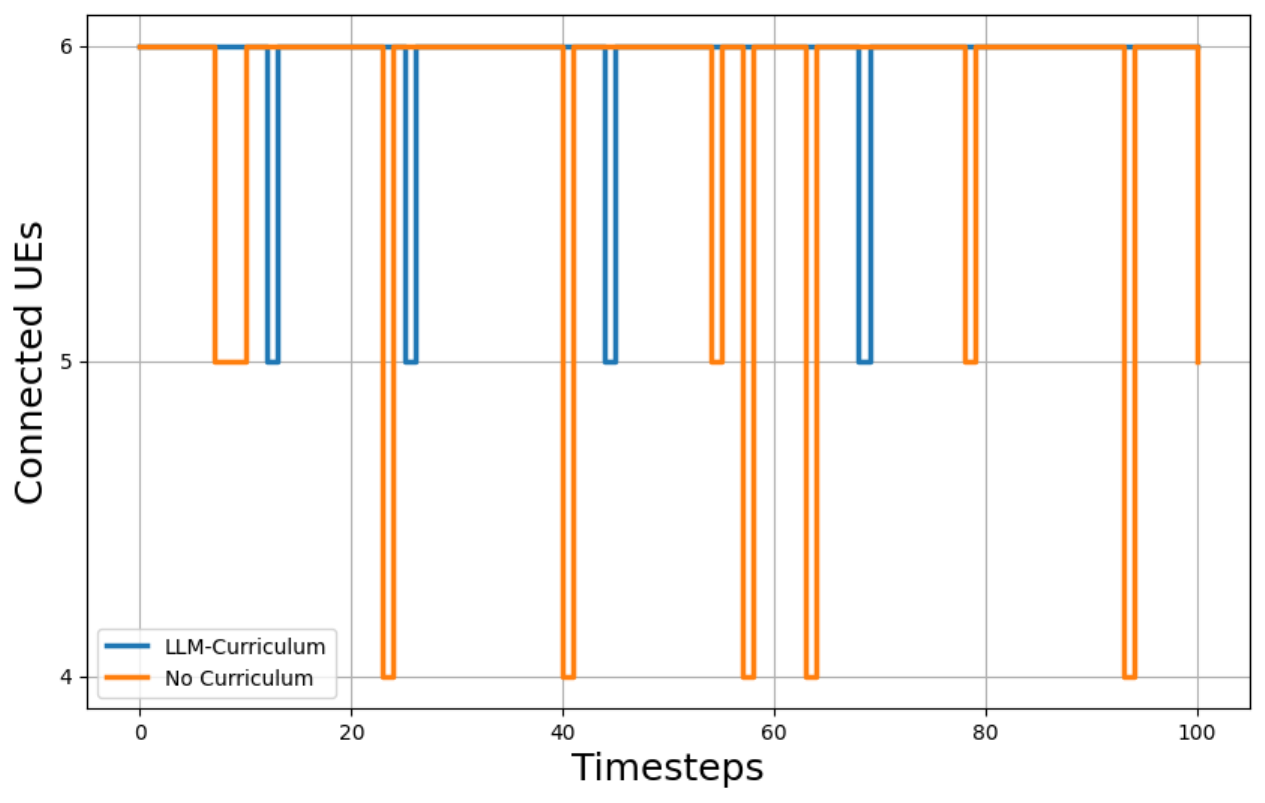}
    \caption{Comparison of connected UEs in an unseen environment with 6 UEs and 3 BSs}
    \label{fig:result3}
\end{figure}

Figure \ref{fig:result2} depicts the average QoE for the proposed approach and the baseline approach as the number of UEs ($M$) increases. The curriculum-trained RL agent demonstrates consistently superior generalization capabilities. Additionally, as $M$ increases, the gap between the two approaches widens. This experiment was repeated multiple times to assess the consistency of the results. Figure \ref{fig:result3} demonstrates the number of connected UEs per step for each agent in a single experiment. The curriculum-based approach exhibits fewer overall UE dropouts compared to the agent trained without a curriculum. This improved performance is likely due to explicitly training the agent on maintaining connections during mobility, as outlined in stage 2 of the curriculum.


\section{Conclusions and Future Direction}

This work introduces a novel framework that leverages LLMs to automatically generate curricula for training RL agents in the context of mobile networks. Through extensive simulations, we demonstrate that the LLM-generated curricula accelerate the convergence of RL agents. Moreover, the curriculum-trained agents exhibit superior generalization capabilities when tested in scenarios beyond their training conditions, such as higher numbers of users.
The key advantage of the proposed framework is mitigating the typically arduous and domain expertise-intensive process of manually designing curricula for RL agents. By automating this process using LLMs' reasoning and generative capabilities, this work opens up new opportunities for more efficiently deploying RL solutions for emergent and complex network management problems. For future generation networks, such autonomous approaches will be crucial for handling the increasing scale, complexity and dynamism. 
While this work focuses on a mobile networking use case, namely user association, the general framework of using LLMs for automating curriculum generation can likely extend to many other RL problem domains that suffer from convergence and generalization challenges. Future research can investigate other use cases in computer and communication networks. Moreover, future research can explore more sophisticated reward and environment modeling approaches, and tighter integration between the LLM and RL components for an end-to-end curriculum generation pipeline. Nevertheless, this work takes an important first step towards a new paradigm of leveraging state-of-the-art language models to supercharge RL capabilities.

\bibliographystyle{IEEEtran}

\bibliography{references}





\end{document}